\documentclass{elsart}  

\usepackage[english]{babel}
\usepackage[utf8x]{inputenc}
\usepackage{amsmath}
\usepackage{graphicx}
\usepackage[textsize=tiny,draft,colorinlistoftodos, textwidth=1.45cm]{todonotes}
\usepackage[hyphens]{url}
\usepackage{soul}
\usepackage{paralist}
\usepackage{smartdiagram}
\usepackage[numbers,sort&compress]{natbib}

\begin{document}
\begin{frontmatter}

\title{A study of existing Ontologies in the IoT-domain}
\author[IIITdelhi]{Garvita Bajaj},
\ead{garvitab@iiitd.ac.in}
\corauth[C]{Corresponding author. Email: garvitab@iiitd.ac.in; }
\author[inria]{Rachit Agarwal},
\ead{rachit.agarwal@inria.fr}
\corauth[C]{Corresponding author. Email: rachit.agarwal@inria.fr;}
\author[IIITdelhi]{Pushpendra Singh},
\ead{psingh@iiitd.ac.in}
\author[inria]{Nikolaos Georgantas},
\ead{nikolaos.georgantas@inria.fr}
\author[inria]{Valerie Issarny}
\ead{valerie.issarny@inria.fr}

\address[IIITdelhi]{Indraprastha Institute of Information Technology, New Delhi, India}
\address[inria]{Inria, Paris, France}

\medskip

\begin{abstract}
Several domains have adopted the increasing use of IoT-based devices to collect sensor data for generating abstractions and perceptions of the real world. This sensor data is multi-modal and heterogeneous in nature. This heterogeneity induces interoperability issues while developing cross-domain applications, thereby restricting the possibility of reusing sensor data to develop new applications. As a solution to this, semantic approaches have been proposed in the literature to tackle problems related to interoperability of sensor data. Several ontologies have been proposed to handle different aspects of IoT-based sensor data collection, ranging from discovering the IoT sensors for data collection to applying reasoning on the collected sensor data for drawing inferences. In this paper, we survey these existing semantic ontologies to provide an overview of the recent developments in this field. We highlight the fundamental ontological \emph{concepts} (e.g., sensor-capabilities and context-awareness) required for an IoT-based application, and survey the existing ontologies which include these concepts. Based on our study, we also identify the shortcomings of currently available ontologies, which serves as a stepping stone to state the need for a common unified ontology for the IoT domain.
\end{abstract}
\end{frontmatter}
\section{Introduction}
With the rapid adoption of the Internet of Things (IoT) technology in various domains including health, transportation, and manufacturing, the number of IoT devices in the world is expected to increase to 50 billion by the end of 2020\footnote{\url{https://www.statista.com/statistics/471264/iot-number-of-connected-devices-worldwide/}}. These IoT devices collect an enormous amount of data using the sensors embedded in them. According to a Cisco report, the amount of annual global data traffic will reach 10.4 ZB (zettabytes) by 2019\footnote{\url{http://www.cisco.com/c/dam/en/us/solutions/collateral/service-provider/global-cloud-index-gci/white-paper-c11-738085.pdf}}. It is interesting to note here that this data will be multi-modal in nature comprising of various formats including video streams, images, and strings. 

Handling such large-scale heterogeneous data and processing it in real-time will be a key factor towards building \emph{smart} applications~\cite{Aggarwal2013}. Semantic approaches - ontologies - have been used to solve these issues related to large-scale heterogeneity. Ontologies are defined as a ``\emph{well-founded mechanism for the representation and exchange of structured information}''~\cite{ye2007ontology}. Existing works have proposed the use of \emph{unified} ontologies to tackle issues of interoperability and automation associated with heterogeneity of sensor data~\cite{atzori2010internet,gyrard2014enrich,nambi2014unified}. However, multiple possible unifications developed by domain experts~\cite{gyrard2014domain} pose several challenges as every unified ontology proposes its self-defined taxonomy.

\begin{figure}
\includegraphics[width=\columnwidth]{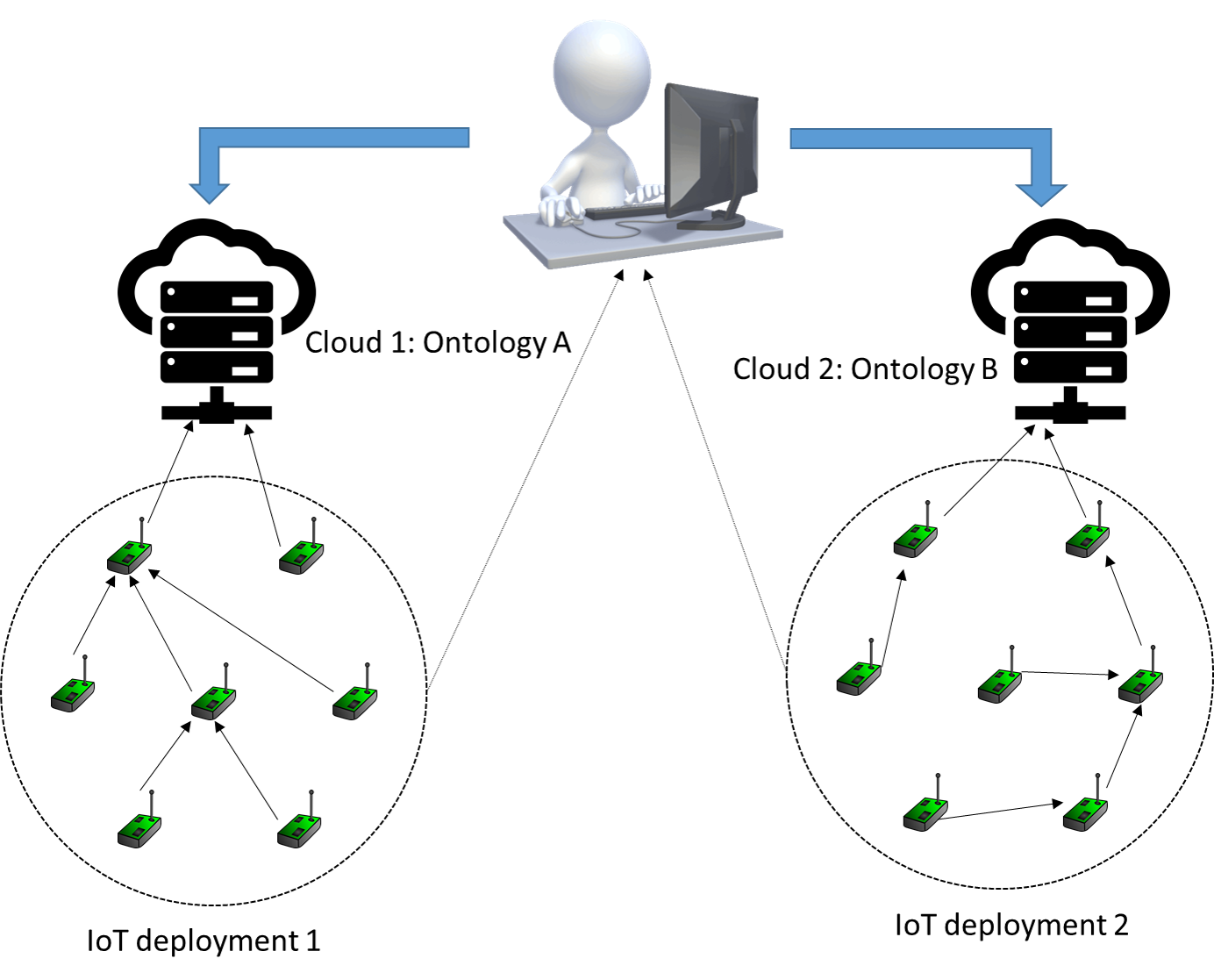}
\caption{Challenges may arise with the use of semantic approaches in IoT applications. The lack of a common semantics may lead to interoperability and design issues.}
\label{fig:heterogeneiry_issues}
\end{figure}

Figure~\ref{fig:heterogeneiry_issues} illustrates an example scenario where the use of several possible unifications of ontologies causes issues. Let us consider two IoT platforms deployed at different geographic locations for building two different applications for smart-health and smart-building domains respectively. Accessing data from both the platforms at another location (e.g., a remote server using data from both the platforms for developing another application) will lead to the challenges described below.

Firstly, the integration of multi-modal sensor data from different sources (i.e., sensors belonging to the two platforms in this scenario) results in complexity and variability in information exchange as multiple sources (i.e., IoT sensors) corresponding to the same sensor data follow the nomenclature~\cite{ye2007ontology} defined by different ontologies. For example, Cloud 1 may store a sensor's location using a tag \texttt{loc} (defined by ontology A); while Cloud 2 may use the tag \texttt{location} for referencing sensor location (defined by ontology B). This difference in nomenclatures may lead to usability issues in developing cross-domain applications as developers would require prior knowledge of every data source and its associated ontology before selecting one for use.

Secondly, the heterogeneous data representation also poses challenges in the development of cross-domain applications that rely on the same set of sensors~\cite{gyrard2014enrich} available with different IoT platforms. For example, the temperature sensor used in the smart-health application and the smart-building application would yield two temperature values - body temperature and room temperature respectively - that cannot be used interchangeably even if the same nomenclature (e.g., \texttt{temperature} or \texttt{Temp}) is used to describe the sensor data.

Thirdly, heterogeneity also causes problems in the design, interaction, and integration of automated solutions based on sensor data~\cite{Dibowski:2011:ODD:1945420.1945423}. For example, with the growing complexity of building automation systems and available devices, there is a need to develop automated design approaches for building systems. Developing such automated approaches would require a comprehensive specification of the hardware and software components to avoid heterogeneity issues, which can be solved using a common unified ontology.

\par 

Defining one comprehensive unified ontology for the domain of IoT may be challenging as there are more than 200 domain ontologies available~\cite{amelieLov4IoT}. For most of the ontologies, there are certain concepts peculiar to the IoT application domains while some concepts used are common to all the IoT platforms. Consider, for example, the two platforms shown in Figure~\ref{fig:heterogeneiry_issues}. Platform 1 will require concepts from health domain while Platform 2 may need concepts from energy and location domains to cater to application needs. On the other hand, both these platforms will require some concepts to answer the basic queries posed by a developer including the location of a sensor and capabilities of a sensor among others. Concepts belonging to the former group (application-specific) comprise the vertical silos of an IoT ontology, while those belonging to the latter group comprise the horizontal silos as shown in Figure~\ref{fig:structureOfOntology}. In this work, we limit our discussion to the unification of concepts pertaining to the horizontal silos in the IoT domain.

In 2012, the W3C Semantic Sensor Networks Incubator Group (SSN-XG)\footnote{\url{https://www.w3.org/2005/Incubator/ssn/ssnx/ssn}} was formed to address the issues of heterogeneity in sensor networks. Although a standard, SSN\footnote{A new version of SSN is available, however, it still does not address the issues discussed.} still fails to address several aspects of the IoT data such as real-time data collection issues, providing a taxonomy for measurement units, context, quantity kinds (the phenomena sensed), and exposing sensors to services, which compels developers to define new concepts in IoT domain while developing novel applications. Several ontologies have been proposed to overcome the shortcomings of SSN, and most of them are limited to certain concepts owing to the vastness of the domain.

In this work, we identify these standard concepts by identifying the \textbf{competency questions} - ``\emph{what are the queries that experts will submit to a knowledge base to find answers?}''~\cite{yan2015ontology}. We identify these questions using the 4W1H methodology~\cite{4w1hDescribing,4w1hcrowdsensing}. The answers to the competency questions will help us identify the core concepts for a unified IoT ontology. In this paper, we survey the existing IoT and IoT-related ontologies proposed since 2012\footnote{For a detailed study on context-aware ontologies in the domain of IoT before 2012, we refer the readers to~\cite{perera2014context}.} to identify the concepts that could be used towards building the unified IoT ontology. We aim to identify these concepts to solve the heterogeneity issue arising because of several domain ontologies available in the literature. From the concepts identified, we also aim to structure the existing literature which can serve as a tool for ontology developers. To summarise, we identify the main concepts in an IoT-based application using the 4W1H methodology and classify the existing ontologies on the basis of these concepts. Our study aims to highlight the research gaps in existing ontologies which will help identify concepts required for a unified \emph{standard} ontology in this domain.
\subsection*{Roadmap}
We begin with identifying the core concepts by defining the competency questions using the 4W1H methodology in Section~\ref{sec:requirements}. We identify a structure within the proposed ontologies with a survey in Section~\ref{sec:survey}. In Section~\ref{sec:evaluation}, we discuss the different approaches used to evaluate the ontologies in order to identify a common ground for comparing existing ontologies with each other. We end this paper with a discussion in Section~\ref{sec:discussion} on the need of a common unified ontology for achieving semantic interoperability among IoT applications.

\section{Identifying Core Concepts}\label{sec:requirements}
\begin{figure}
\centering
 \includegraphics[width=0.6\columnwidth]{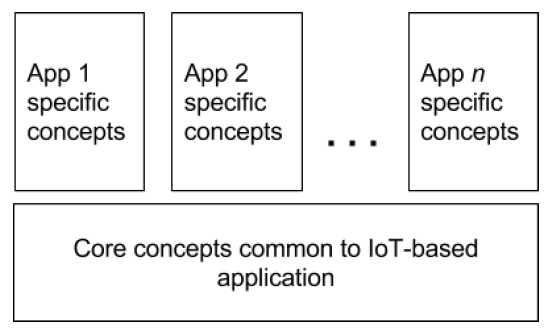}
 \caption{Distributing the concepts requirement of an IoT-based application into horizontal and vertical elements}
 \label{fig:structureOfOntology}
 \end{figure}
An application based on IoT technology (e.g., a standalone mobile application or a cloud-based application using static sensors) consists of several components. These components are used to interact with several different sensors that are attached to a platform. These sensors may be statically positioned or dynamically moving for collecting data across a geographical area of interest. An ideal IoT ontology should describe the core concepts common to all IoT applications (i.e., horizontal), and concepts that are specific to applications (i.e., vertical) as shown in Figure~\ref{fig:structureOfOntology}. However, to limit the concepts in a unified ontology for IoT domain, we focus on including all the horizontal core concepts in this study. To identify these fundamental concepts required in an ontology, we must identify the \textit{competency questions} - queries submitted by experts to a knowledge base to find answers~\cite{yan2015ontology}. Previous works by Haller et al.~\cite{haller2013domain}, Wang et al.~\cite{wang2013knowledge}, and Bermudez-Edo et al.~\cite{bermudez2016iot,BermudezEdo2017} have identified some of these questions to list the following different core concepts for the IoT domain: \texttt{Augmented Entity}, \texttt{User}, \texttt{Device}, \texttt{Resource}, and \texttt{Service} among others. These studies (\cite{haller2013domain,wang2013knowledge,bermudez2016iot,BermudezEdo2017}), however, do not consider all the core concepts which can be used by several applications to infer complex information in IoT scenarios. 
\par 4W1H~\cite{4w1hDescribing,4w1hcrowdsensing} is a popular approach that is being used to describe the basics of an event/situation. To define the comprehensive list of concepts for an IoT ontology, we rely on the use of 4W1H methodology. For this, we must define five variables - four \textbf{W}s (\textbf{W}hat, \textbf{W}here, \textbf{W}hen, \textbf{W}ho) and one \textbf{H} (\textbf{H}ow) to list the competency questions required for identification of the core IoT concepts. These variables and the corresponding competency questions are listed below followed by a detailed explanation of the identified concepts: 
\begin{itemize}
\item{\textbf{Who}: \textit{Who} will provide the information required to develop the IoT application? The answer to this question requires concepts to identify the sources which provide data for building an IoT-based application. This source is the \textit{\textbf{sensor}} which is embedded in a \textit{\textbf{platform}} that can be a part of a \textit{\textbf{testbed}}. Thus, to answer the `who' question, an IoT ontology must include concepts for sensor, platform, and testbed.} 
\item{\textbf{What:} \textit{What} are the conditions under which the source must collect data? Should the data be collected under certain circumstances only? Answering this question requires the IoT ontology to include concepts to define the \textit{\textbf{context}} of the data source (e.g., the mobility of the sensor, the activity being performed, whether the measurement was taken by the device automatically or was there some human intervention).}
\item{\textbf{Where:} \textit{Where} should the data come from? This refers to the \textit{\textbf{location}} of the data source which can be defined using geo-coordinates, building names, landmarks, etc. The IoT ontology must incorporate the different ways to locate the data source for allowing developers to identify the data source.}
\item{\textbf{When: }\textit{When} should the data collection happen? Should the sensor collect data at specified \textit{\textbf{timestamps}}, should it be at a regular frequency, or over a span of time? An IoT ontology should provide concepts to support different formats for defining the time for data collection.}
\item{\textbf{How: }Once the data has been collected by the relevant sensors, \textit{how} should it be exposed to the developer for building the IoT application? There should be concepts to support \textit{\textbf{services}} for providing developers with the access to this sensor data.}
\end{itemize}
These competency questions help us identify the requirements of the core concepts in a unified IoT ontology. We list and explain these concepts in detail below (the concepts are highlighted in bold):

\begin{enumerate}
\item An IoT application is based on sensor data collected from various heterogeneous sensors. A \textbf{sensor} is defined as a ``\emph{source that produces a value representing a quality of a phenomenon}''~\cite{compton2009survey}. Note that here sensor data refers to both raw data captured by the sensor and the metadata that describes a sensor.
\item These sensors need a power supply to operate. They may be battery-powered or may be attached to another power source for energy. This ``\emph{identifiable entity to which a sensor is attached}'' is called a \textbf{platform}~\cite{compton2009survey}.
\item For a wide-scale deployment of IoT solutions provided by these platforms, a large-scale realistic \textbf{testbed} is required~\cite{gluhak2011survey}. This testbed should provide functionalities to support the different kinds of sensors required (based on the phenomenon they sense), and should provide mechanisms to communicate the sensor data to the applications. The testbeds may also internally store the captured sensor data in their proprietary format for performing local data analysis. 
\item Given a testbed infrastructure (e.g., SmartSantander\footnote{\url{http://www.smartsantander.eu/}}, IoT-Lab\footnote{\url{https://www.iot-lab.info/}}, Ambiciti\footnote{\url{https://io.ambiciti.mobi}}), an identifiable \textbf{service} to access this raw or processed data is required. The service should support a combination of sensor data to get better contextual information (assimilation), and removal of redundant sensor data (filtering) from multiple data sources. The service should allow the users to discover information from the environment based on certain search criteria. 
\item Further, since most of the sensor data is \textbf{location} and \textbf{time specific}, the location and time information may be used to provide a common \textbf{context} (``information about a location, its environmental phenomena, and the people, devices, objects, and software agents it contains''~\cite{chen2003ontology}) for modeling the services in IoT frameworks~\cite{flury2004owl}).
\end{enumerate}
    
Several domains have proposed ontologies defined by domain experts for these concepts. In this work, we study some ontologies centered around these concepts.

\section{Survey}\label{sec:survey}
In this section, we present a survey of ontologies in the IoT-domain based on the concepts identified above. These ontologies are prevalent in several application areas including Building Management Systems (BMS)~\cite{balaji2016brick}, indoor navigation~\cite{szasz2016iloc}, and smart-homes~\cite{Bae201432}. Ye et al.~\cite{ye2007ontology} have proposed a classification of ontologies based on (i) \emph{expressiveness} as light-weight or heavy-weight; or (ii) \emph{generality} as generic, domain-specific, or application-specific. In this paper, we use the latter classification based on the functionalities provided by the ontology. Since our focus is on the entire IoT domain and not on a specific application, we restrict our discussion to generic and domain specific ontologies only.

\subsection{Sensor ontologies}
\begin{figure}[t]
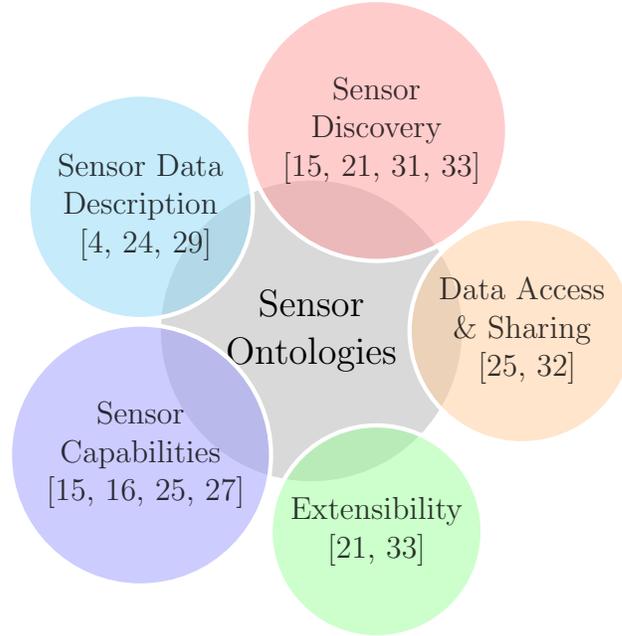

\centering
\smartdiagram[bubble diagram]{Sensor\\Ontologies, Sensor\\Discovery\\~\cite{Hirmer2016,saref,balaji2016brick,bermudez2016iot}, Sensor Data\\Description\\~\cite{ssn2012,nachabe2015unified,gyrard2014enrich}, Sensor\\Capabilities\\~\cite{ontosensor,xue2015ontology,bermudez2016iot,BermudezEdo2017}, Extensibility\\\cite{balaji2016brick,saref}, Data Access \\\& Sharing\\~\cite{xue2015ontology,shi2012sensor}}
\caption{The different problem areas tackled by existing sensor ontologies}
\label{fig:sensorOntologies}
\end{figure}

Existing sensor ontologies proposed in the literature aim to solve heterogeneity problems associated with the hardware, software, and the data management aspect of sensors. These aspects are highlighted in Figure~\ref{fig:sensorOntologies} where we classify the different sensor-based ontologies based on the problems they tackle. Please note that `sensor data' may refer to the raw sensor data generated by sensing the phenomenon, or the metadata information used to describe the sensor capabilities. Here, we use the term `sensor data' to refer to the raw data generated by sensors, and `sensor capabilities' to describe the metadata information (e.g., the accuracy of a sensor and coverage range of a sensor.) associated with these sensors. In the following text, we briefly introduce the different proposed ontologies based on the generic concepts defined by them.

\subsubsection{Generic Ontologies}
In 2012, W3C (World Wide Web Consortium) proposed a standard ontology - \textbf{SSN} (Semantic Sensor Network) ontology~\cite{ssn2012} for describing the sensor \texttt{resources} and the data collected through these sensors as \texttt{observations}. SSN ontology aims to solve the heterogeneity problems associated with \textit{sensor discovery} and \textit{sensor data collection} but has limited concepts to support the spatial and temporal association of sensor data with the \texttt{resources}. Also, it is difficult to describe the different sensor capabilities using SSN as there is only one concept for \texttt{sensor description}. To overcome this problem, \textbf{Xue et al.}~\cite{xue2015ontology} proposed an ontology with concepts for sensor types - \texttt{normal} or \texttt{advanced}, and also sensor capabilities - \texttt{static}, or \texttt{dynamic}. They also introduce concepts to deal with issues of sensor management and data sharing in sensor networks. However, their ontology supports a small number of sensors in terms of the phenomenon they can sense, and provides semantic support for only a limited number of sensor features. Further, in the ontology, the concept of location is only limited to building rooms and building floors. This limits the usage of this ontology to indoor applications only. Gyrard et al.~\cite{gyrard2014enrich,gyrard2014standardizing} overcome this limitation by providing \textbf{M3} ontology - a comprehensive ontology proposed as an extension to W3C's SSN ontology to support the description of sensors, observations, phenomena, their units, and domains, which also allows for reasoning on sensor data using rules to infer contextual information.
 
Since the platform to which the sensors are attached can be mobile, issues with dynamicity and sensor discovery must also be dealt with. Concepts for dealing with these issues were introduced in \textbf{OntoSensor}~\cite{ontosensor}. The ontology extends concepts from SensorML\footnote{\url{https://en.wikipedia.org/wiki/SensorML}}, ISO-19115\footnote{\url{http://bit.ly/2nedvM1}}, and SUMO~\cite{niles2001towards}, to allow concepts for the identification of sensor categories, behavior, relationships, functionalities, and meta-data regarding sensor characteristics, performance, and reliability. OntoSensor aims to support interoperability and ontology-based inferences that require aspects of physical sensing to be incorporated in the ontology definition. The ontology is heavy-weight, has high usage complexity, and lacks the ability to describe sensor \texttt{observations}. This limitation was solved by \textbf{MyOntoSens}~\cite{nachabe2015unified} which provides a generic and exhaustive ontology for describing sensor \texttt{observations} and capabilities to reason over the collected data. MyOntoSens has been proposed for the domain of wireless sensor networks (WSN) and borrows several concepts and relationships from existing ontologies including OntoSensor~\cite{ontosensor}, SSN~\cite{ssn2012}, and QUDT~\cite{hodgson2014qudt} which makes it applicable to the IoT domain also. MyOntoSens provides concepts to support sensor discovery and sensor registration with the platform.

\textbf{Hirmer et al.}~\cite{Hirmer2016} propose an ontology to support dynamic registration and bindings of new sensors to a platform. They borrow the concepts of \texttt{Sensor}s, \texttt{Thing}s and their associated properties from SensorML, and introduce an additional concept of `\texttt{Adapter}' associated with every sensor. The borrowed concepts support sensor discovery while the newly introduced concepts provide/compute additional information about sensor data. For example, \texttt{adapter}s are used to compute the average quality of sensor data values from the quality of the sensor provided by the manufacturer and the staleness of the values generated by the sensor. This additional information is then used to build a repository of the different possible sensor bindings - bindings of different \texttt{Sensor}s with different \texttt{Adapter}s - which is then used to automate the registration and binding tasks. 

\textbf{Shi et al.}~\cite{shi2012sensor} identified the problems associated with inconsistencies in concept definitions among existing ontologies and proposed a framework to overcome them. Their ontology, \textbf{Sensor Core Ontology} (SCO), borrows concepts from existing sensor ontologies with provision to add new concepts, thereby supporting extensibility. The ontology focused mainly on concepts related to sensor data where each sensor observation is associated with time, space (location), and a theme (phenomenon being sensed). However, the description of these concepts themselves is not detailed enough even though the focus is on aspects of sensor data collection. 

\subsubsection{Domain-specific ontologies}
In this section, we present the domain specific ontologies and limit our discussion to domains including smart appliances, energy, and building management systems which also form an integral part of the IoT ecosystem. Note that this is not the comprehensive list of all the domains. There are 17 different domains that have been identified\footnote{\url{http://sensormeasurement.appspot.com/?p=ontologies}}, but we focus only on the above-mentioned domains.
\textbf{SAREF}~\cite{saref} (Smart Appliance REFerence) ontology exists in the domain of \textit{smart appliances} and aims to reuse and align concepts and relationships in existing appliance-based ontologies. The concept of \texttt{functions} - one or more commands supported by \texttt{devices} (sensors) defined in SAREF - supports modularity and extensibility of the ontology, and also helps in maintenance of the appliances. An extension of SAREF - \textbf{SAREF4EE}~\cite{daniele2016interoperability} has been proposed to support interoperability with EEBus\footnote{\url{https://www.eebus.org/en/about-us/}} and Energy@Home\footnote{\url{http://www.energy-home.it/SitePages/Home.aspx}} standards. Along similar lines, \textbf{Dey et al.}~\cite{dey2014sensor} propose an extension of OntoSensor ontology in the energy domain to include the spatial and temporal concepts of sensor data. Another ontology which focuses on energy efficiency at building level - \textbf{Brick}~\cite{balaji2016brick}, has been proposed in the domain of \textit{Building Management Systems} (BMS). Brick proposes the use of \textit{tags} and \textit{tagsets} to specify sensors and building subsystems, thus, enabling sensor discovery. The use of \textit{tags}, \textit{tagsets}, and \textit{functional block}s provides support for extensibility and flexibility for developing cross-building BMS applications. They do not cover the entire range of sensors available in the market and limit their focus to sensors used in smart-building applications. Since the concept of \texttt{location} used by them is limited to `zones' (HVAC zones, rooms, and floors) inside the buildings, it can't be extended to other application areas.  

\subsection{Context-Aware ontologies}
As defined by \textbf{Chen et al.}~\cite{chen2003ontology}, contexts are ``\textit{used to describe places, agents, and events}''. Contexts can be classified as external or internal; and physical or logical~\cite{baldauf2007survey}. Several ontologies have been proposed for context-aware systems to effectively label contextual information collected from sensor devices in the form of sensor data. The following section reviews some of the generic and domain specific context ontologies. We emphasize more on domain specific ontologies as the context inferred in context-aware applications (including IoT-based applications) is highly dependent on the domain in consideration. As an example, for labeling a user activity, we need a complete understanding of the possible activities in the domain as activities performed on a university campus will be very different than the activities performed in a smart-home environment. Thus, better concepts for semantic labeling of contexts can be identified with domain-specific ontologies.
\subsubsection{Generic Ontologies}
In IoT applications, contexts are often important to correctly interpret sensor data~\cite{perera2014context}. Thus, as the first step towards semantic labeling of contexts in IoT applications, \textbf{Baldauf et al.}~\cite{baldauf2007survey} propose the common architecture principles of context-aware systems based on the classification of contexts: external (physical) or internal (logical). External or physical contexts are those that can be measured using physical sensors, while internal or logical contexts are those that are explicitly specified by users or captured by monitoring user interactions (e.g., a user's goal or emotional state). The authors also derive a layered conceptual design framework to explain the elements common to most context-aware architectures. To further describe the concepts for contexts (\texttt{places}, \texttt{agents}, and \texttt{events}), \textbf{Chen et al.}~\cite{chen2003ontology} propose \textbf{COBRA-ONT} for smart spaces. They describe places using $<$\texttt{lat}, \texttt{lon}, \texttt{string-name}$>$ with two kinds of places with different constraints. There are also \texttt{agents} such as human and software, which are located in places and play certain roles to perform some activities. An important contribution is the broker architecture proposed by the authors, which can be used to acquire and reason over contextual information from mobile devices in order to reduce the burden of developers.

Since in IoT applications, the concept of a location may vary from a \textit{point} to a \textit{place of interest}, COBRA-ONT is one of the most promising ontologies to represent the location context of ``things''. It not only can describe a point but also be used to specify a place using a `string' value for a location.
To further correlate contextual information with location, \textbf{Kim et al.}~\cite{locationReasoning} propose an ontology based on the information provided by mobile device sensors, both - physical (e.g. WiFi, Bluetooth, etc.) and virtual (e.g., user schedule, web-logs, etc.)  to support context-aware services. The proposed ontology defines the relations between different user locations and the contexts identified. The authors further propose a reasoner upon this ontology and evaluate it to identify locations. The result shows that their reasoner has higher location accuracy than the GPS locations.

\subsubsection{Domain-specific ontologies}
Several domain-specific ontologies have been proposed for defining contextual information as the contexts inferred from sensor data may vary largely based on the domain. For example, in a university domain, the activities may be limited to only a small subset such as reading, playing, sleeping, etc., while in a smart-home, a large set of activities (e.g., cooking, cleaning, reading, sleeping, etc.) may be encompassed. In this section, we will limit our study to context-aware ontologies in the domain of smart-homes and activity recognition.
 
Smart-home technologies are now being developed to assist disabled and elderly individuals with dignified living. \textbf{Okeyo et al.}~\cite{Okeyo2014155} propose an ontology to semantically label the \textbf{Activities of Daily Living} (ADL) such as cooking food, brushing teeth, etc. for the smart-homes domain. Their ontology is based on dynamic segmentation of sensor data for variable time windows to identify simple user activities. These simple activities are then used to infer more complex activities. A limitation of their ontology is that it focuses only on individual activities, and does not consider social activities (e.g., tea party, business meetings, etc.) that might occur in a smart-home environment. To accommodate these concepts, Bae et al.~\cite{Bae201432} present \textbf{RADL} (Recognizing Activities of Daily Living) system to classify three different kinds of services in smart-home environments. The ontology supports the discovery of devices and their location in smart-home domains using concepts like \texttt{Person}, \texttt{Sensor}, \texttt{Device}, \texttt{Location}, and \texttt{AmIApplication}. Reasoning on this ontology helps the system discover the user activities such as individual and social.

Further, extending our study to activity recognition in the domain of a university, Lee et al.~\cite{lee2017location} propose \textbf{UAO} (University Activity Ontology) which caters to activities specific to a university campus (e.g.: attending a lecture or having lunch in the cafeteria). The authors use concepts defined in \textbf{CONON} (CONtext ONtology)~\cite{wang2004ontology} and introduce new concepts that are specific to activities on a university campus. They propose a hierarchical division of concepts using sub-ontologies to differentiate existing concepts from the new concepts. The concepts borrowed from existing ontologies are placed in the upper hierarchical structure, while the newly introduced concepts occupy positions in the lower hierarchical structure.

\subsection{Location-based ontologies} Location is used to describe the spatial context (partly, physical context) of users/devices. Although a subset of context, we consider location ontologies separately as they have been used in several different areas like defining cultural heritage~\cite{cacciotti2013monument}, urban planning~\cite{liao2015place}, etc., beyond IoT technologies. However, we limit our discussion to location ontologies defined in the domain of IoT only.

\subsubsection{Generic Ontologies}
\textbf{WGS84} ontology~\cite{brickley2006basic} describes abstract concepts for defining \texttt{SpatialThing}s such as buildings, people, etc., and \texttt{TemporalThing}s such as events, or time durations. It also describes the geographical locations of these `things' by using concepts for defining the geo-coordinates using \texttt{latitude}, \texttt{longitude}, and \texttt{altitude}. Concepts defined in this ontology are inherited from abstract concepts for defining subclasses specific to the system. A more descriptive location ontology is provided by \textbf{Flury et al.}\cite{flury2004owl} for context-aware services as they identify the location to be a common denominator for modeling services in context-aware environments. 
They provide a generic ontology for the location concept for ``device-based services encountered in ubiquitous computing environments''. They provide abstract mathematical models to categorize the different location solutions (like Cartesian coordinates and numerical estimation techniques) used to define location information. The different models considered are as follows: (i) \textit{Geometric models} (comprising of Cartesian coordinates); (ii) \textit{Set-theoretic models} (for defining location as an element of a set, e.g., Cellular location, WiFi AP location, etc.); (iii) \textit{Graph-based models} (for defining locations in physically grounded networks, social networks, etc.), and (iv) \textit{Semantic models} (for defining locations defined using human-friendly notations). However, in most IoT-based applications, location data is often collected using sensor data. \textbf{Kim et al.}~\cite{kim2015ontology} rely on the use of sensor data collected from mobile devices of users to estimate device location. They propose a reasoner built upon sensor data and information from location sensors to estimate the location of users.
\subsubsection{Domain-specific Ontologies}
In this section, we restrict our discussion to location ontologies for indoor navigation.
\par 
\textbf{iLoc}~\cite{szasz2016iloc} is an ontology specified in the domain of indoor building navigation and follows some of the best practices for defining a new ontology. It uses concepts borrowed from several existing ontologies - \texttt{QUDT}\footnote{\url{http://www.qudt.org/}}, W3C Geo vocabulary, and \texttt{vCard}\footnote{\url{https://www.w3.org/TR/vcard-rdf/}} and also supports extensibility. Since the root concept \texttt{Location} is borrowed from W3C Geo vocabulary, iLoc can also be extended to provide navigation for outdoor locations. The authors, however, do not provide an extensive evaluation of the ontology, and only demonstrate the usability through a use-case scenario.
\subsection{Time-based ontologies}
Time is used to describe the temporal context. Most of the IoT-related ontologies reuse existing ontologies that define the temporal context. 
\subsubsection{Generic Ontologies}
Many ontologies that define \textbf{temporal} context have been proposed in the literature, for example, \textbf{DAML-Time}\footnote{\url{http://www.cs.rochester.edu/~ferguson/daml/}} (DARPA Agent Markup Language project Time initiative), \textbf{DAML-S}\footnote{\url{http://www.daml.org/services/daml-s/0.9/}} (DAML for Web Services), \textbf{KSL-Time}\footnote{\url{http://www.ksl.stanford.edu/ontologies/time}} (Stanford Knowledge Systems Lab Time ontology~\cite{Fikes}) and \textbf{OWL-Time ontology}\footnote{\url{http://www.w3.org/2006/time\#}} to name a few. DAML-Time focused on concepts to provide a common understanding of time. DAML-S however, provides temporal concepts required to define a web service such as profile, process and time. We refrain ourselves from describing them in detail. KSL-Time ontology, on the other hand, provides concepts to distinguish between different types of intervals and granularity.

The most commonly used ontology is the OWL-Time ontology proposed by Hobbs and Pan~\cite{hobbsOwlTime2002} that focuses on describing date-time information specified in Gregorian calendar format. Several updates to the initial version of OWL-Time ontology have been proposed and implemented. The latest version of OWL-Time ontology\footnote{\url{http://w3c.github.io/sdw/time/}} provides more concepts while deprecating some. In general, the OWL-Time ontology provides concepts to describe time using five main core concepts, namely, \texttt{TemporalEntity}, \texttt{Instant}, \texttt{Interval}, \texttt{ProperInterval}, and \texttt{DateTimeInterval}. M. Gr\"uninger in~\cite{Gruninger} verified the older version of the OWL-Time ontology in an independent study. A light-weight version of the OWL-Time ontology called \textbf{Time-Entity} is developed for applications needing only limited concepts~\cite{PanTimeEntry}. 

\textbf{Timeline} ontology\footnote{\url{http://motools.sourceforge.net/timeline/timeline.html#term_Instant}} is yet another time-based ontology. It extends Time Ontology by providing various concepts such as \texttt{ContinuousTimeLine}, \texttt{DiscreteTimeLine}, \texttt{OriginMap}, etc. \textbf{Fikes et al.} in~\cite{Fikes} considered granularity of time to provide more flexibility to the annotations. They considered both open and closed intervals to include flexibility on top of temporal aspects such as \texttt{Time-Point}. 

\subsubsection{Domain-specific Ontologies}

\textbf{TimeML}, a time Domain-Specific Language (DSL), has been proposed by Pustejovsky et al.~\cite{Pustejovsky} to provide the ability to query specific data based on duration, events, granularity and instant. An instance of TimeML is defined in XML format. However, Pustejovsky et al. used TimeML to annotate events within a Natural Language Processing system. \textbf{Dey et al.}~\cite{dey} extended the use of TimeML to IoT systems and as a use-case, demonstrated the usage of TimeML to enrich annotations for the Energy sensors. \textbf{Zhang et al.} in~\cite{ZhangChinese} created a time ontology that is capable of representing temporal aspects: \begin{inparaenum} \item characterized by events either cultural or historical; and \item in Chinese calendar instead of Gregorian calendar.\end{inparaenum}

\section{Ontology Evaluation}\label{sec:evaluation}
The goal of an ontology is to provide a semantic framework which can solve the problem of heterogeneity and interoperability associated with domain applications. This requires available ontologies to be comprehensive, readable, extensible, scalable, and reusable. Several methods have been proposed to evaluate an ontology on these different aspects. As a first step, validation tools like \textbf{OntoCheck}~\cite{schober2012ontocheck} and \textbf{SAOPY}\footnote{\url{http://iot.ee.surrey.ac.uk/citypulse/ontologies/sao/saopy.html}} have been developed. These validation tools help ontology creators to identify inconsistencies in naming conventions and metadata completeness for (e.g., cardinality checks, class hierarchy, etc.). Another tool called \textbf{OOPS}\footnote{\url{http://oops.linkeddata.es}} was developed to identify pitfalls in the ontology~\cite{oops} based on 41 evaluation criteria\footnote{\url{http://oops.linkeddata.es/catalogue.jsp}} such as missing annotations, missing domain or range, and identifying anomalies in the relationships. \textbf{Vapour}\footnote{\url{http://linkeddata.uriburner.com:8000/vapour}} is yet another validation tool that focuses on identifying if the ontology is correctly published or not based on the best practices recipes\footnote{\url{https://www.w3.org/TR/swbp-vocab-pub/}}. \textbf{RDF Triple Checker}\footnote{\url{http://graphite.ecs.soton.ac.uk/checker/}} is another tool that validates triples by finding typos and matching namespaces for the used concepts. \textbf{Bae et al.}~\cite{Bae201432} have also proposed a schema-based approach for evaluating the design of ontologies on their richness, depth, width, and inheritance. Note that there are other validation tools also available\footnote{A comprehensive list of existing semantic validators can be found on \url{https://www.w3.org/2001/sw/wiki/SWValidators}}, but we restrict ourselves from comparing them with each other.
\par 
Once these basic checks have been performed, ontologies must be evaluated on parameters like scalability and extensibility. \textbf{Bermudez et al.}~\cite{bermudez2016iot,BermudezEdo2017} evaluate IoT-lite on scalability and complexity using Round Trip Time (RTT) as a measure. They annotate sensor data using their ontology and execute SPARQL queries on annotated datasets containing a different number of triples. For evaluating ontologies on extensibility, \textbf{Shi et al.}~\cite{shi2012sensor} propose a metric for objectively assessing the nomenclature of concepts used in the ontology based on translatability, clarity, comprehensiveness, and popularity.
\par
A survey by \textbf{Brank et al.}~\cite{brank2005survey} highlights the following four major approaches to evaluate an ontology: \begin{inparaenum} \item by comparing it with another ontology in the same domain; \item by comparing the concepts identified with a set of documents containing information about the domain in consideration; \item by proposing an application using the ontology to justify its usage; and \item by asking domain experts to evaluate it manually\end{inparaenum}. Summarizing our study, we observe that no single metric or approach can be used to evaluate an ontology. This has also been identified by the authors in~\cite{vrandevcic2009ontology} where they identify eight different parameters to assess the quality of an ontology. These are accuracy, adaptability, clarity, completeness, computational efficiency, conciseness, consistency, and organizational fitness. They also develop a framework called Semantic MetaWiki (SMW) to evaluate ontologies on these parameters. They identified that no single parameter could be used to compare ontologies; the goodness of an ontology depends on the application in consideration, and it cannot be judged by a single metric. 

Most of the works referenced in this survey have demonstrated the feasibility of their ontologies using sample applications which are only a measure of the usability of the ontology. For evaluating ontologies on other parameters, there are independent measures but no single comprehensive metric for concrete assessment. 

\section{Discussion}\label{sec:discussion}
Often systems propose proprietary ontologies for achieving semantic consistency in different modules. However, ontologies proposed by one system can often be used by other systems to improve interoperability. For example, authors in~\cite{desai2015semantic} use W3C's SSN ontology and propose a system to provide semantic interoperability between the different protocols used by different classes of IoT applications such as wearable devices, and hardware platforms like Raspberry Pi, Arduino, etc. Another example of such a system is \textbf{STAR-CITY} (\textbf{S}emantic \textbf{T}raffic \textbf{A}nalysis and \textbf{R}easoning for \textbf{CITY})~\cite{lecue2014star}. This system uses multiple domain ontologies\footnote{\url{http://goo.gl/5TbTT2
}} to combine heterogeneous data sources - static and dynamic, generating data that has variety, has high sampling rate, and volume. The system provides traffic analysis and reasoning for efficient urban planning which supports re-usability of heterogeneous, real-time data, and expose them to the end users through REST-ful web-APIs by reusing and combining existing domain ontologies instead of proposing a new one.
\par 
Another approach to reuse existing concepts in ontologies, while adding new concepts specific to the domain, is to provide a hierarchical structure composed of two sub-ontologies as suggested by \textbf{Lee et al}.~\cite{lee2017location}. The authors use existing concepts as part of the `upper' ontology and define new domain-specific concepts as part of the `lower' ontology in the hierarchy of sub-ontologies. This hierarchical division enhances the readability, and thus improves the scope of reusability of the proposed ontology.
\par
Several works have now started reusing existing ontologies by combining them to propose new ontologies for the IoT domain. These new ontologies are meant for specific platforms aimed at collecting and integrating IoT data. However, they lack one or more concepts identified in Section~\ref{sec:requirements}, rendering them incomplete for the IoT domain. \textbf{IoT-Lite}~\cite{bermudez2016iot}, for example, provides a lightweight instantiation of the SSN ontology and extends it using SAO. They also include dynamic semantics by introducing mathematical formulas to estimate missing sensor values during the data annotation phase itself. This saves time as the data is no longer required to be sent to a server for extrapolating missing sensor values, rendering IoT-Lite dynamic, fast, and interoperable; thus, making it suitable for constrained IoT environments. The concepts covered by IoT-Lite include sensor information and location. \textbf{VITAL} ontology~\cite{kazmi2016overcoming} also follows a similar approach by combining concepts from SSN, QUDT, OWL-Time, and WGS84 to define sensors, measurements, time, and location concepts respectively. \textbf{OpenIoT} ontology~\cite{vanderSchaaf2015} also uses SSN as a base to build upon concepts required for IoT applications and testbeds - \texttt{Observation}, \texttt{Sensor}, and \texttt{Location}. It further extends to define utility metrics for Service level agreements between its users and provided OpenIoT services. 

Authors of \textbf{IoT-O} ontology~\cite{Seydoux2016} claimed to built IoT-O ontology such that it is modular and focused on two sets of requirements - \textit{Conceptual} and \textit{Functional}. They defined ``conceptual" requirements as those requirements that form the core of any IoT-related ontology. These were based on the description of devices, data, services and their lifecycle. Further, ``functional" requirements were defined as those requirements that follow best practices defined by the semantic community. IoT-O is one of the first approaches towards unification of the IoT ontologies. IoT-O reuses concepts from SSN, SAN, DUL, QUDT, oneM2M and further, defines new concepts. Nevertheless, the ontology lacks a complete/holistic view of context. \textbf{FIESTA-IoT} ontology~\cite{agarwal2016unified} is yet another attempt to unify existing ontologies for the IoT domain. It covers most of the concepts identified in Section~\ref{sec:requirements}. However, it lacks concepts for annotating context information such as \textit{place of interest}, has no support for context-aware services and has limited notion for activity, actuators, virtual entities, etc. \textbf{oneM2M} ontology~\cite{oneM2M2016}, on the other hand, although supported by standardization bodies in IoT, also lacks contextual information. oneM2M provides minimal concepts needed for representing a device and its functionality. oneM2M does not reuse any existing ontology to define the concepts. Another ontology called \textbf{Open-MultiNet} (ONM)~\cite{Willner2015} is built to define federations of infrastructures. Similar to oneM2M, ONM does not reuse concepts. Further, ONM also lacks the notion of context. 

An early attempt by \textbf{Hachem et al.}~\cite{hachem2011ontologies} to solve the challenges of heterogeneity and scalability in IoT, proposed three ontologies for a service-oriented middleware for IoT applications. They are - (i) \textit{Device ontology} (for describing the physical things and their metadata), (ii) \textit{Domain ontology} (for defining relations between physical objects and mathematical formulas/functions), and (iii) \textit{Estimation ontology} (for describing automated estimation processes to relate one mathematical function with another, and to model the errors in estimations). These ontologies lack the concepts required to annotate the collected data for addressing the issues of heterogeneity and interoperability. A detailed model to define concepts for devices, services, context, and information is also provided by \textbf{IoT-A} ontology~\cite{bauer2013iot} for the IoT Architecture Reference Model~\cite{bassi2013enabling}. Note that, as we have described earlier, all these ontologies are designed to satisfy specific requirements for the platform they are applied in and thus, do not completely follow the requirements set in Section~\ref{sec:requirements}.
\par
Even with so many ontologies defined for IoT-specific platforms, a single comprehensive ontology comprising of the core concepts is missing. Developers should, therefore, reuse existing ontologies and merge them into new ontologies as per their requirements. Before defining a new ontology, they should identify the competency questions. If the competency questions can be answered by existing ontologies, then developers should emphasize on reusing concepts and relationships from existing ontologies. This allows easy integration of heterogeneous data and facilitates developing reasoning capabilities on top of semantic data. The reasoners can provide useful insights about the systems and their output/reasoning data can also serve as raw data for other systems/applications. However, if a developer has to define new concepts for his/her systems, he/she should follow the best practices. These best practices are: reusing existing concepts as much as possible and aligning the concepts with equivalence relations~\cite{Suarez2010}, validating the ontology using ontology validators (see Section \ref{sec:evaluation}), having recommended ontology metadata\footnote{\url{https://w3id.org/widoco/bestPractices}}, making the ontology accessible, following a modular approach so that the ontology is reusable, and documenting it with samples to demonstrate its usage~\cite{gyrard2015semantic}.
\section{Conclusion}
In this paper, we study existing ontologies in the domain of IoT and identify the major concepts listed by them. We identify that no existing ontology is comprehensive enough to document all the concepts required for semantically annotating an end-to-end IoT application as ontologies are often restricted to a certain domain. Hence, as a step towards developing a comprehensive unified ontology for IoT domain, we identify the core concepts required for developing an IoT application. We study the ontologies proposed for those concepts and analyse the methods used for their evaluation. Our study identifies that there are no concrete methods for evaluating an ontology, and developers must always follow the best practices while publishing a new ontology in order to enhance readability, usability, extensibility, and interoperability.
\section*{Acknowledgement}
This work is partially funded by the European project ``Federated Interoperable Semantic IoT/cloud Testbeds and Applications (FIESTA-IoT)'' from the European Union's Horizon 2020 Programme with the Grant Agreement No. CNECT-ICT-643943.
\bibliographystyle{elsart-num}
\bibliography{references.bib}
\end{document}